\documentclass[letterpaper, 10 pt, conference]{ieeeconf}
\IEEEoverridecommandlockouts
\usepackage{xcolor}
\usepackage{float}
\usepackage{multirow}
\usepackage{float}
\usepackage{graphicx} 
\usepackage{booktabs}
\usepackage{multirow}
\usepackage{pifont}
\usepackage{amsmath}  
\usepackage{amssymb}
\usepackage[noadjust]{cite}
\usepackage{hyperref}
\usepackage[normalem]{ulem}
\usepackage{bold-extra}   
\usepackage{colortbl}

\newcommand{\papername}{AnyViewDex}
\newcommand{\depthyes}{\cellcolor{red!20}Yes}
\newcommand{\depthno}{\cellcolor{green!12}No}
\newcommand{\bestRGB}[1]{\cellcolor{green!20}\best{#1}}
\newcommand{\bestDepth}[1]{\cellcolor{yellow!40}\best{#1}}

\title{\LARGE \bf
\papername{}: View-Invariant Dexterous Manipulation from RGB Observations
}

\author{Soham Patil$^{1*}$, Om Sanjay Gunjal$^{1*}$, Sourabh Bhosale$^{1}$, Arhan Chavare$^{2}$, Ramandeep Singh Hora$^{3}$ and Spandan Roy$^{1}$%
\thanks{*These authors contributed equally to this work.}%
\thanks{$^{1}$Soham Patil, Om Sanjay Gunjal, Sourabh Bhosale and Spandan Roy are with the Robotics Research Center, IIIT-H, India.
        {\tt\small soham.patil@research.iiit.ac.in}
        {\tt\small omgunjalmtt@gmail.com}
        \newline {\tt\small sourabhrbhosale03@gmail.com}
        {\tt\small spandan.roy@iiit.ac.in}}%
\thanks{$^{2}$Arhan Chavare is with the Department of Electronics Engineering,
        Veermata Jijabai Technological Institute Mumbai, India
        \newline {\tt\small aachavare\_b24@et.vjti.ac.in}}%
\thanks{$^{3}$Ramandeep Singh Hora is with the Department of Electrical Engineering and Computer Sciences,
        Indian Institute of Science Education and Research Bhopal, India
        \newline {\tt\small ramandeep23@iiserb.ac.in}}%
}

\newcommand{\best}[1]{\textbf{\boldmath #1}}
\newcommand{\second}[1]{\underline{#1}}

\newcommand{\etal}{\textit{et al.}}

\makeatletter
\let\@oldmaketitle\@maketitle
\renewcommand{\@maketitle}{%
    \@oldmaketitle

    \vspace{-0.4cm}
    \begin{center}
        \includegraphics[width=\linewidth]{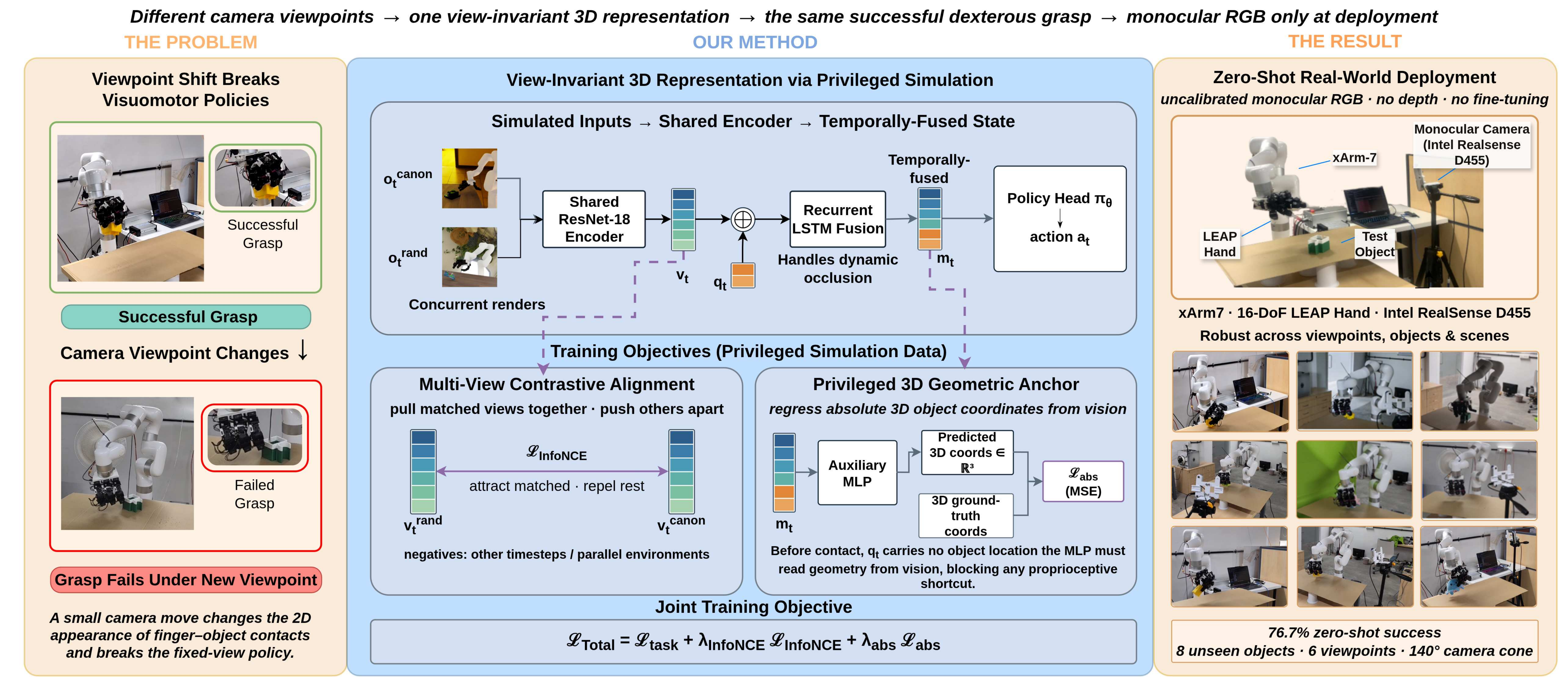}

        \refstepcounter{figure}
        \@makecaption{Figure~\thefigure}{%
            \textbf{\papername{} Overview.} 
\textbf{(Left)} Camera shifts alter 2D contact geometry, causing fixed-view policies to fail. 
\textbf{(Center)} In simulation, privileged 3D coordinate regression ($\mathcal{L}_{\text{abs}}$) grounds multi-view contrastive embeddings ($\mathcal{L}_{\text{InfoNCE}}$) in metric space. 
\textbf{(Right)} Discarding privileged targets at deployment, the policy executes zero-shot dexterous grasping on a 16-DoF LEAP Hand using solely uncalibrated monocular RGB and proprioception (76.7\% success across 8 unseen objects and 6 viewpoints).%
        }
        \label{fig:teaser}
    \end{center}

    \vspace{-0.4cm}
}
\makeatother
 
\begin{document}
\maketitle
\setcounter{figure}{1}
\thispagestyle{empty}
\pagestyle{empty}

\begin{abstract}
Visuomotor policies for multi-fingered dexterous manipulation are highly sensitive to camera viewpoint shifts. To achieve view invariance, recent methods increasingly rely on explicit 3D modalities like RGB-D or point clouds, which can introduce hardware dependencies, calibration requirements, and vulnerability to sensor noise during real-world deployment. In this work, we show that view-invariant control can be achieved without explicit test-time 3D sensing by encoding geometric knowledge into the visual representation during simulation. We present \papername{}, an asymmetric training pipeline that combines multi-view contrastive alignment with privileged 3D geometric supervision. By regressing absolute 3D object coordinates during simulated training, this auxiliary objective provides a geometric grounding signal that mitigates the spatial collapse of the globally pooled contrastive embedding. At deployment, the policy operates zero-shot using only uncalibrated monocular RGB and proprioception. We validate this approach across both reinforcement learning and student-teacher distillation. In hardware evaluation on an xArm7 with a 16-DoF LEAP Hand, \papername{} reaches 76.7\% grasping success across eight unseen objects and six uncalibrated viewpoints (480 trials; 2{,}400 across all ablation conditions), indicating that geometrically grounded monocular policies transfer zero-shot without test-time depth. Project Page: https://anyviewdex.github.io/
\end{abstract}

\section{Introduction}

\label{sec:introduction}
Visuomotor policies enable robots to learn dexterous manipulation directly from images \cite{qin2023dexpoint,mandlekar2023mimicgen,chen2023visual}, but they often overfit to fixed camera viewpoints. The problem is acute in dexterous manipulation: unlike parallel-jaw grippers, multi-fingered hands coordinate numerous joints to establish precise contacts \cite{andrychowicz2020learning}, and when the camera shifts, the 2D appearance of these contacts changes, creating spatial ambiguity when inferring 3D finger-object relationships from an uncalibrated 2D image.

To resolve these geometric ambiguities, recent methods increasingly rely on explicit 3D modalities like RGB-D or point clouds \cite{qin2023dexpoint,yuan2024learning,li2026manivid}. While these approaches naturally handle viewpoint changes, they depend on test-time depth sensors, which can introduce hardware dependencies, minimum-distance clipping, and vulnerability to sensor noise during deployment \cite{keselman2017intel}. Alternatively, 2D Spatial Transformer Networks \cite{yang2023movie} struggle with large out-of-plane 3D rotations. Multi-view contrastive objectives \cite{pang2025learning,seo2023multi} encourage broad view invariance, but global contrastive pooling optimizes for semantic similarity and discards the local geometric detail required for fine dexterous control~\cite{9578497,Xie_2021_CVPR}.

In this work, we show that view-invariant control can be achieved without test-time 3D sensing by embedding geometric knowledge into the visual representation during simulation. Contrastive alignment enforces cross-view correspondence but lacks metric scale and absolute grounding. \papername{} resolves this by jointly optimizing an InfoNCE multi-view loss with privileged regression of absolute 3D object coordinates ($\mathcal{L}_{\text{abs}}$), anchoring the globally pooled embedding in metric space across uncalibrated viewpoints. Confining all privileged supervision to simulation, this asymmetric strategy enables zero-shot deployment from monocular RGB and proprioception.

We evaluate \papername{} across reinforcement learning (RL) and student-teacher distillation. In simulation, \papername{} attains the highest success among RGB-only methods on three of the four Maniwhere tasks and exceeds the RGB-D Maniwhere baseline on \textit{Close Dex} (closing a hinged laptop lid; 92.1\% vs.\ 81.5\%), while trailing RGB-D and spatial-grid RGB baselines on fine localization of a small object (\textit{Lift Cube Dex}, 53.0\%). On physical hardware (an xArm7 with a 16-DoF LEAP Hand), \papername{} reaches 76.7\% zero-shot success over 480 trials across eight unseen objects and six uncalibrated viewpoints, against 30.2\% for an identically trained domain-randomization baseline under matched conditions.

Following prior view-generalization work \cite{yuan2024learning, yang2023movie, pang2025learning, seo2023multi}, we use view-invariant to mean insensitive to uncalibrated camera placement within a training distribution (here, a $140^\circ$ azimuthal cone), without per-view calibration or fine-tuning.

Our contributions are: 
(1) \papername{}, an asymmetric representation learning framework that combines multi-view contrastive alignment with privileged 3D geometric supervision to mitigate spatial collapse in monocular RGB control.
(2) Cross-paradigm validation demonstrating that this geometrically grounded representation supports continuous control in both end-to-end RL and student-teacher distillation. 
(3) Sim-to-real hardware evaluation (2{,}400 trials across all conditions) of zero-shot view-invariant grasping, reaching 76.7\% success without test-time depth or real-world fine-tuning.
\section{Related Work}
\label{sec:related_work}
To achieve view invariance in visuomotor policies, research has increasingly shifted from brute-force data scaling toward structured representation learning, spanning generative view synthesis, masked reconstruction, spatial warping, and explicit 3D feature alignment.

\paragraph{Data Scaling and Generative View Synthesis.}
A common approach to viewpoint invariance is training on data collected from diverse camera poses. While large-scale datasets improve robustness \cite{tian2024view, walke2023bridgedata}, acquiring dense multi-view demonstrations for high-DoF dexterous manipulation remains expensive \cite{wang2024dexcap}. Recent works address this through latent diffusion-based novel view synthesis to augment single-view datasets \cite{tian2024view,chen2024rovi} . Although these methods reduce the need for extensive multi-view data and calibration, synthesis artifacts can still degrade the geometric precision required for dexterous grasping.

\paragraph{Reconstruction-Based Representations.}
Other frameworks learn robust features by training encoders to reconstruct scenes from masked or novel viewpoints. Masked World Models (MWM) \cite{seo2023masked} and its multi-view extension, MV-MWM \cite{seo2023multi}, utilize masked autoencoders to reconstruct target views from masked source images during training. While this formulation extracts single-view representations at inference time, pixel-level reconstruction prioritizes photometric detail over the geometric and contact cues essential for dexterous manipulation.

\paragraph{Spatial Adaptation and 2D Feature Alignment.}
Rather than relying on reconstruction, an alternative paradigm dynamically adapts to viewpoint shifts through spatial warping or explicit feature disentanglement. Methods like MoVie \cite{yang2023movie} tackle this by incorporating Spatial Transformer Networks (STNs) to actively warp visual features into a canonical frame at test time, mathematically compensating for camera displacement. Similarly, to explicitly separate task-critical information from camera-specific artifacts, methods like ReViWo \cite{pang2025learning} utilize contrastive losses on 2D images to push the network to learn disentangled view-independent and view-dependent features. 

\paragraph{3D Representation Learning.}
Recent methods leverage explicit 3D information to enforce spatial consistency. Maniwhere \cite{yuan2024learning} combines RGB-D observations with contrastive learning, while ManiVid-3D \cite{li2026manivid} operates directly on point clouds to learn view-invariant representations. Although 3D sensing naturally handles viewpoint changes, it remains vulnerable to depth noise, missing measurements, calibration errors, and self-occlusions. Contemporaneously, Jiang \etal{} \cite{jiang2025you} condition policies on explicit camera extrinsics via Pl\"ucker ray-maps; this reinforces the value of geometric grounding but requires strict test-time calibration, whereas \papername{} shifts all spatial reasoning to simulation and needs no geometric input at deployment.

\paragraph{Privileged Representation Learning.}
To combine 3D spatial grounding with the deployment simplicity of monocular vision, we leverage privileged representation learning, drawing on asymmetric actor-critic formulations where privileged state stabilizes optimization. \papername{} applies this to view-invariant representation learning, using privileged 3D data only in simulation to anchor a contrastive RGB backbone, avoiding generative hallucination, photometric reconstruction, and test-time depth.
\section{Methodology}
\label{sec:method}

Our core insight is that a lightweight, globally pooled 1D visual embedding can acquire the metric spatial awareness of depth-reliant systems when geometric knowledge is instilled during simulation via a privileged 3D auxiliary objective.

\subsection{Monocular Visual Encoding and Contrastive Alignment}
\label{subsec:visual_encoding}

\begin{figure}[h]
    \centering
    \includegraphics[width=0.475\textwidth]{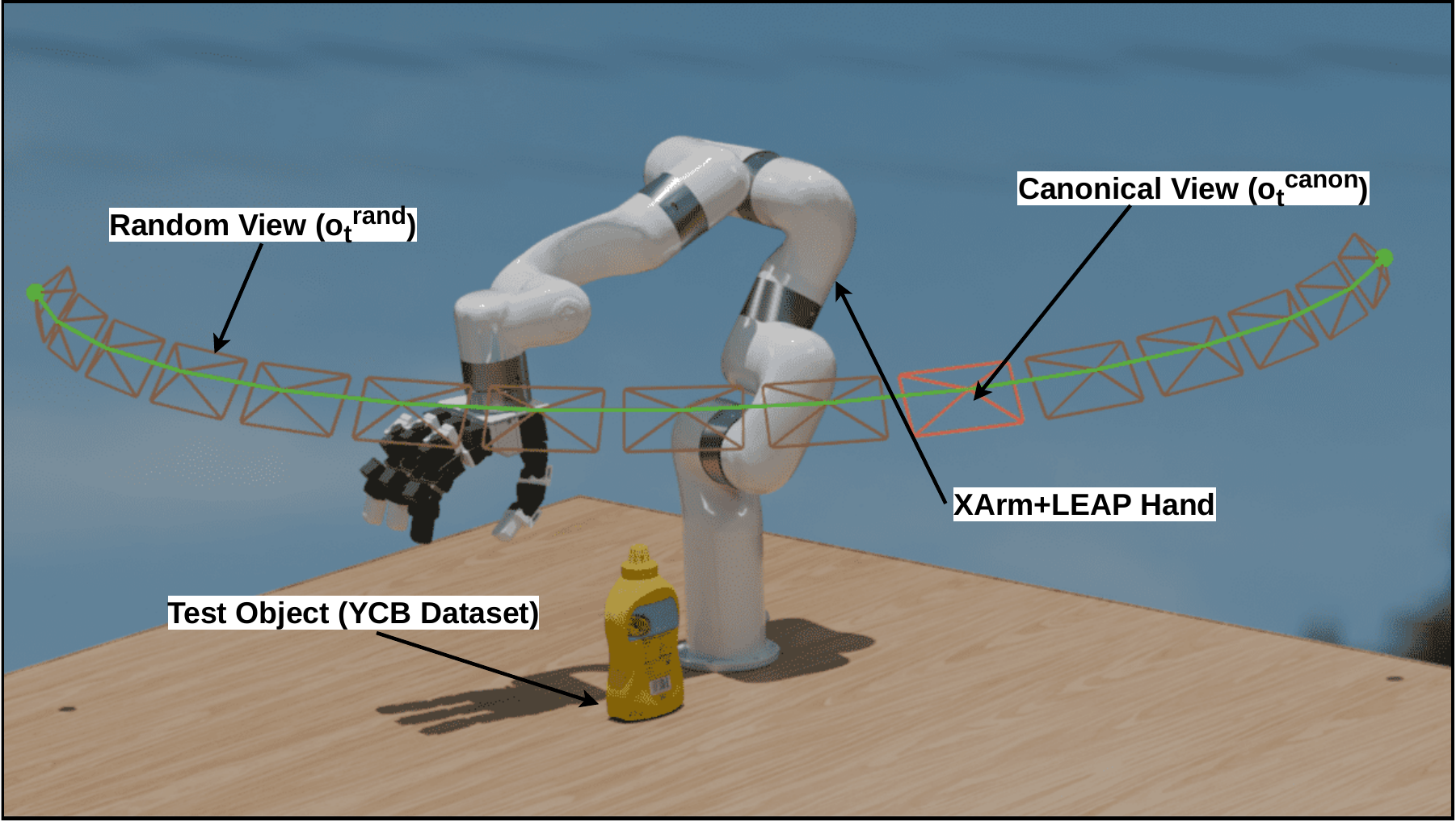}
    \caption{\textbf{Multi-View Contrastive Training Setup.} During simulated representation learning, visual observations of the identical physical state are rendered concurrently.}
    \label{fig:sim_multi_view}
\end{figure}

At each timestep $t$, the environment generates a visual observation $\mathbf{o}_t \in \mathbb{R}^{3 \times H \times W}$. Rather than utilizing computationally heavy Vision Foundation Models or high-dimensional spatial feature maps, we employ a lightweight ResNet-18 encoder \cite{he2016deep}, applying global average pooling to the final convolutional layer to project the image into a compact 1D visual embedding $\mathbf{v}_t \in \mathbb{R}^{d_v}$ suitable for high-frequency continuous control.
\begin{table*}[t]
\centering
\footnotesize
\renewcommand{\arraystretch}{0.88}   
\small
\setlength{\tabcolsep}{11pt}         
\caption{%
  \textbf{Calibration-Free Viewpoint Robustness: Maniwhere RL Suite}
  (mean success rate\,\% $\pm$ std., 5 seeds; $\uparrow$ higher is better).
  \colorbox{green!20}{\textbf{Green}}: best RGB-only per task;
  \colorbox{yellow!40}{\textbf{Yellow}}: depth oracle (DO);
  \underline{underlined}: second-best RGB-only.
}
\label{table:sim-results}
\begin{tabular}{clccccc}
\toprule
  & \multirow{2}{*}{\textbf{Method}}
  & \multirow{2}{*}{\textbf{Depth?}}
  & \multicolumn{4}{c}{\textbf{Task Success Rate (\%)} $\uparrow$} \\
\cmidrule(lr){4-7}
  & & & \textbf{Lift Cube Dex} & \textbf{Pick \& Place Dex}
      & \textbf{Close Dex} & \textbf{Button Dex} \\
\midrule
\multirow{6}{*}{\rotatebox{90}{\tiny\textsc{RGB Prior Work}}}
  & MV-MWM~\cite{seo2023multi}             & \depthno
    & \bestRGB{$78.0 \pm 5.1$}   & \second{$34.0 \pm 28.9$}
    & \second{$69.5 \pm 19.7$}   & $77.6 \pm 14.3$ \\
  & SGQN~\cite{bertoin2022look}             & \depthno
    & $14.0 \pm 7.7$              & $3.2  \pm 4.6$
    & $15.0 \pm 5.8$              & $12.8 \pm 4.4$ \\
  & SRM~\cite{huang2022spectrum}            & \depthno
    & $24.4 \pm 8.0$              & $6.4  \pm 5.9$
    & $\phantom{0}8.0 \pm 4.2$   & $18.8 \pm 4.1$ \\
  & MoVie~\cite{yang2023movie}              & \depthno
    & $\phantom{0}6.0 \pm 2.2$   & $1.0  \pm 2.2$
    & $\phantom{0}6.0 \pm 4.1$   & $11.3 \pm 4.7$ \\
  & PIE-G~\cite{yuan2022pre}               & \depthno
    & $10.5 \pm 2.2$              & $1.0  \pm 2.3$
    & $\phantom{0}5.0 \pm 3.5$   & $11.3 \pm 4.7$ \\
  & Maniwhere~(RGB)~\cite{yuan2024learning} & \depthno
    & \second{$72.5 \pm 1.5$}    & $0.0  \pm 0.0$
    & $17.3 \pm 2.7$              & \second{$82.4 \pm 9.6$} \\
\midrule
\rotatebox{90}{\tiny\textsc{D.O.}}
  & Maniwhere~(RGB-D)~\cite{yuan2024learning} & \depthyes
    & \bestDepth{$88.8 \pm 8.9$} & \bestDepth{$76.4 \pm 9.2$}
    & \bestDepth{$81.5 \pm 5.6$} & \bestDepth{$97.6 \pm 1.2$} \\
\midrule
\multirow{4}{*}{\rotatebox{90}{\tiny\textsc{Ablations}}}
  & Fixed Camera   & \depthno
    & $0.0  \pm 0.0$   & $0.0  \pm 0.0$
    & $0.0  \pm 0.0$   & $0.0  \pm 0.0$ \\
  & DR Only        & \depthno
    & $26.2 \pm 3.6$  & $3.5  \pm 0.2$
    & $0.0  \pm 0.0$   & $24.0 \pm 3.2$ \\
  & w/o Aux.\ Loss & \depthno
    & $0.8  \pm 0.1$   & $0.0  \pm 0.0$
    & $0.0  \pm 0.0$   & $88.0 \pm 5.0$ \\
  & w/o InfoNCE    & \depthno
    & $22.0 \pm 7.0$   & $33.2 \pm 3.4$
    & $22.0 \pm 10.0$  & $35.2 \pm 5.0$ \\
\midrule\midrule
\rotatebox{90}{\tiny\textsc{Ours}}
  & \textbf{\papername{}~(Ours)} & \depthno
    & $53.0 \pm 5.0$
    & \bestRGB{$72.4 \pm 3.6$}
    & \bestRGB{$92.1 \pm 5.9$}
    & \bestRGB{$96.0 \pm 2.0$} \\
\bottomrule
\end{tabular}
\vspace{-0.4cm}
\end{table*}
Global pooling discards the explicit 2D spatial layout of the scene, and a low-resolution feature grid would retain some of it. We nonetheless pool to a single vector for two reasons: it yields a compact temporally-aggregated state (an LSTM hidden state or a frame stack) small enough for high-frequency control, and it exposes a single global embedding on which the contrastive and coordinate-regression objectives act without a spatial-correspondence step. The design choice is to discard the grid and reinstate metric geometry through the auxiliary loss (Section~\ref{subsec:geometric_regularization}) rather than carry it in feature maps; \textit{Lift Cube Dex} marks where this costs accuracy (Section~\ref{sec:conclusion}). To retain view-invariant semantic features in the pooled vector, we apply an InfoNCE contrastive objective \cite{oord2018representation} during simulated training. The simulator concurrently renders the identical physical state from a fixed canonical camera ($\mathbf{o}_t^{\text{canon}}$) and a randomized tracking camera ($\mathbf{o}_t^{\text{rand}}$), as shown in Figure \ref{fig:sim_multi_view}. Both images are processed by the shared ResNet backbone to produce embeddings $\mathbf{v}_t^{\text{canon}}$ and $\mathbf{v}_t^{\text{rand}}$, which are projected through an MLP head $h(\cdot)$ for alignment. The downstream policy trunk consumes only the raw, unprojected vector from the randomized tracking camera ($\mathbf{v}_t^{\text{rand}}$); the canonical view is used solely for the contrastive loss and discarded at inference.

We optimize the network to maximize the similarity between these cross-view pairs while repelling embeddings of temporally distinct states using the InfoNCE objective:
\begin{equation}
    \mathcal{L}_{\text{InfoNCE}} = -\log \frac{\exp \left( \frac{\mathbf{v}_t^{\text{canon}} \cdot \mathbf{v}_t^{\text{rand}}}{\tau} \right)}{\sum_{j} \exp \left( \frac{\mathbf{v}_t^{\text{canon}} \cdot \mathbf{v}_j^{\text{rand}}}{\tau} \right)}
\end{equation}
where $\tau=0.1$ is a temperature scaling parameter applied to $L_2$-normalized embeddings, and index $j$ iterates over negative sample pairs from other timesteps or parallel environments within the training batch. This alignment encourages variations in camera extrinsics to map to nearby regions in the latent space, providing the semantic basis for view invariance.

\subsection{Temporal Aggregation for Dynamic Occlusion}
\label{subsec:sensor_fusion}
Continuous finger articulation creates dynamic self-occlusions, so a single frame is often insufficient to localize the target during contact. We aggregate temporal context into a latent state $\mathbf{m}_t$ consumed by both the policy and the auxiliary head, and instantiate the aggregator per paradigm (Section~\ref{subsec:policy_optimization}). For student-teacher distillation, we concatenate the visual vector $\mathbf{v}_t$ with proprioception $\mathbf{q}_t$ (joint positions and velocities) into $\mathbf{x}_t = [\mathbf{v}_t, \mathbf{q}_t]$ and process it with a Long Short-Term Memory (LSTM) network \cite{hochreiter1997long}, whose hidden state carries object position through frames where the hand occludes it. For end-to-end RL, we retain the Maniwhere baseline's frame-stacking: $\mathbf{m}_t$ is the stack of recent visual embeddings, without proprioception. In both settings $\mathbf{m}_t$ denotes the temporally-aggregated state passed downstream.

\subsection{Privileged Geometric Regularization}
\label{subsec:geometric_regularization}

A contrastive objective on a globally pooled embedding can be minimized by matching broad semantic content across views while discarding metric geometry; we term this failure mode spatial collapse. Removing the 2D feature grid through global average pooling increases the risk. The contrastive-only variant (\textit{w/o Aux}, Table~\ref{table:sim-results}) shows it directly, falling to near $0\%$ on three of the four tasks. To retain metric precision we add a privileged 3D auxiliary head.

During simulated training, an auxiliary Multi-Layer Perceptron (MLP) regresses task-relevant 3D spatial states, specifically the absolute world-frame coordinates of the target object $\mathbf{p} \in \mathbb{R}^3$. It operates on the temporally-aggregated latent state $\mathbf{m}_t$ (Section~\ref{subsec:sensor_fusion}), whose composition is paradigm-specific:
\begin{equation}
    \hat{\mathbf{p}}_t = \text{MLP}_{\text{aux}}(\mathbf{m}_t)
\end{equation}

To ground the globally pooled embedding in physical space, we optimize this head via a Mean Squared Error (MSE) objective against the ground-truth coordinates extracted directly from the privileged simulator backend:
\begin{equation}
    \mathcal{L}_{\text{abs}} = \frac{1}{B} \sum_{i=1}^{B} \left\| \hat{\mathbf{p}}^{(i)} - \mathbf{p}^{(i)} \right\|_2^2
\end{equation}
where $B$ denotes the batch size. Regressing absolute world-frame coordinates, rather than relative offsets, requires the pooled embedding to decode to a metric position from every training viewpoint. From uncalibrated views this forces the encoder to recover depth and scale from image cues rather than a view-specific 2D layout, driving the geometry needed for dexterity into the channel dimensions and removing the need for depth sensors or 4D feature maps at test time.

Ultimately, our complete visual representation is learned through the joint optimization objective $\mathcal{L}_{\text{Total}} = \mathcal{L}_{\text{task}} + \lambda_{\text{InfoNCE}} \mathcal{L}_{\text{InfoNCE}} + \lambda_{\text{abs}} \mathcal{L}_{\text{abs}}$, where the auxiliary gradients are backpropagated through the recurrent fusion module and the visual encoder simultaneously with the core policy gradients.

\begin{figure*}[t]
    \centering
    \includegraphics[width=\textwidth]{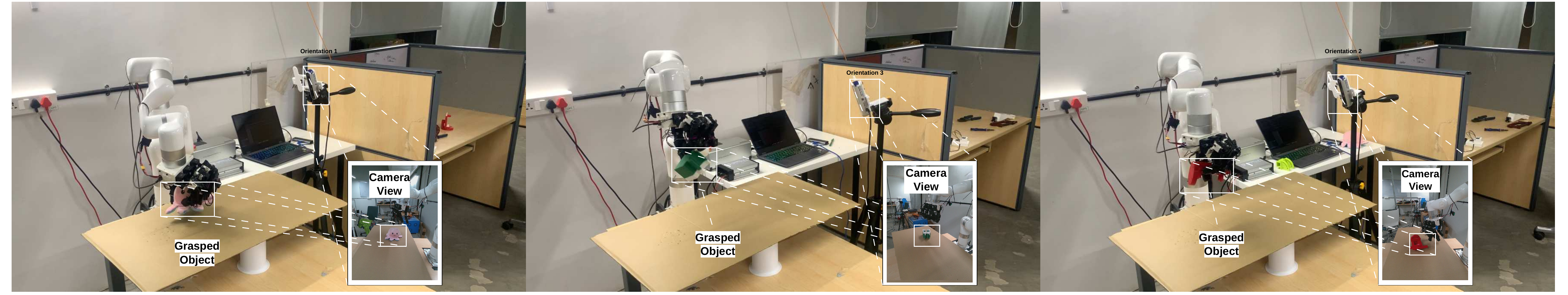}
    \caption{\textbf{Hardware Setup and Zero-Shot View-Invariant Deployment.} xArm7 with a 16-DoF LEAP Hand, observed by a single uncalibrated D455 (RGB only), at three of the six evaluation viewpoints. Insets show the RGB input to the policy.}
    \label{fig:3_panel_deployment}
\end{figure*}
\textbf{Preventing Kinematic Shortcuts:} The proprioceptive shortcut arises only in the distillation configuration, where $\mathbf{m}_t$ fuses $\mathbf{q}_t$: once the hand contacts the object, proprioception could satisfy $\mathcal{L}_{\text{abs}}$ without vision. Two factors preclude this. Episodes start from a fixed neutral arm pose with randomized object locations, so early-trajectory predictions cannot come from kinematics. A linear probe on the frozen visual embedding $\mathbf{v}_t$, with no proprioceptive input, recovers 3D coordinates at $R^2=0.814$ (Section~\ref{subsec:probing}), locating the metric signal in the visual backbone. In the frame-stacked RL configuration $\mathbf{m}_t$ carries no proprioception, so the head regresses object position from visual features by construction.

\textbf{Object Orientation:} The auxiliary head regresses only 3D translation ($\mathbb{R}^3$); orientation is left to the control objective ($\mathcal{L}_{\text{task}}$), keeping the anchor minimal and avoiding over-constraint. Grasp success under uncalibrated views (Tables~\ref{table:sim-results},~\ref{table:ablation-results}) indicates the representation resolves the hand-object alignment that multi-fingered grasping requires.

\subsection{Cross-Paradigm Policy Optimization}
\label{subsec:policy_optimization}

The temporally-aware, geometrically-grounded state $\mathbf{m}_t$ serves as the input for continuous control in both paradigms. To test view invariance independent of the optimization algorithm, \papername{} routes $\mathbf{m}_t$ into a paradigm-specific policy trunk $\pi_\theta$. For asymmetric student-teacher distillation, the policy outputs a continuous stochastic action distribution $\mathcal{N}(\mu, \Sigma)$. For end-to-end RL, the policy outputs a deterministic action vector $\mathbf{a}_t$. In both paradigms, the auxiliary geometric objective grounds the latent space, keeping optimization stable under extreme camera randomization.

\paragraph{Paradigm-Specific Network Architectures}
To ensure fair comparison, we match each paradigm's native control trunk. For Maniwhere (RL) we bypass the explicit LSTM and adopt the baseline's frame-stacking, feeding concatenated visual embeddings to the standard Maniwhere actor and twin-Q critic; for DextrAH (distillation) we fuse the visual embedding with proprioception through an LSTM. The privileged 3D auxiliary head branches from the final temporally-aggregated latent state, regressing the most recent frame's coordinates in the frame-stacked Maniwhere case. Layer dimensions follow the respective baseline trunks and are reported with our released configuration.

\paragraph{Full Training Objective}
The task loss $\mathcal{L}_{\text{task}}$ in the joint objective (Section~\ref{subsec:geometric_regularization}) is paradigm-specific. For Maniwhere (RL), we use the DrQ-v2 algorithm, an off-policy method that optimizes a twin-Q critic via Temporal Difference loss and trains a deterministic actor via the deterministic policy gradient (with fixed exploration noise applied during data collection). For DextrAH (distillation), it is a DAgger behavior-cloning loss formulated as an uncertainty-weighted MSE between the student and privileged-teacher stochastic action distributions, penalizing divergence most heavily along the low-variance (high-confidence) teacher dimensions.

\section{Experiments and Results}

\subsection{Experimental Setup}

\textbf{Hardware Platform and Workspace}
Our 23-DoF platform integrates an xArm7 with a LEAP Hand (Figure~\ref{fig:3_panel_deployment}). The table sits $11.5$~cm above the robot base to maximize downward reach and avoid wrist singularities, an offset mirrored in simulation to prevent a Cartesian sim-to-real gap.

\textbf{Camera Randomization.} Both environments render two concurrent streams per timestep: a fixed canonical view and a randomized extrinsic view resampled each episode. Bounds and distributions are in Table~\ref{tab:camera_randomization}.

\begin{table}[htbp]
\centering
\footnotesize
\renewcommand{\arraystretch}{0.92}
\caption{%
  \textbf{Simulation Camera Randomization.}
  Extrinsic and intrinsic ranges resampled per episode for the randomized
  tracking camera during training; the canonical camera is held fixed.
  All parameters are sampled uniformly.
}
\label{tab:camera_randomization}
\resizebox{\columnwidth}{!}{%
\begin{tabular}{llc}
\toprule
\textbf{Environment} & \textbf{Camera Parameter} & \textbf{Range} \\
\midrule
\multirow{4}{*}{\textbf{Maniwhere}}
  & Orbital Azimuth (Yaw)            & $[-60^\circ,\ 60^\circ]$\textsuperscript{$\dagger$} \\
  & Orbital Elevation (Pitch)        & $[-12.5^\circ,\ 7.5^\circ]$ \\
  & Radial Distance Scaling          & $[0.8\times,\ 1.1\times]$ \\
  & Field of View                    & $[38^\circ,\ 46^\circ]$ \\
\midrule
\multirow{5}{*}{\textbf{DextrAH}}
  & Orbital Azimuth (Yaw)            & $[-110^\circ,\ 30^\circ]$ \\
  & Vertical Elevation Offset        & $[-0.15,\ 0.15]$~m \\
  & Orbital Pitch Jitter             & $[-3^\circ,\ 3^\circ]$ \\
  & Cartesian Origin Noise ($X,Y,Z$) & $[-0.03,\ 0.03]$~m \\
  & Base Rotation Noise (Roll, Pitch, Yaw) & $[-3^\circ,\ 3^\circ]$ \\
\bottomrule
\end{tabular}%
}

\parbox{\columnwidth}{\footnotesize\textsuperscript{$\dagger$}\,\textit{Close Dex} uses an offset azimuthal span of $[0^\circ,\,120^\circ]$.}
\end{table}

\textbf{Simulated Environments.} We benchmark \papername{} across two paradigms: the MuJoCo-based Maniwhere suite \cite{yuan2024learning} for end-to-end RL, and an Isaac Lab environment \cite{mittal2025isaac} inspired by DextrAH \cite{singh2025dextrah} for asymmetric student-teacher distillation via DAgger \cite{ross2011reduction}.
\textbf{Real-World Hardware.} Deployment uses the same xArm7 and LEAP Hand~\cite{shaw2023leap}, observed by a single uncalibrated Intel RealSense D455 (monocular RGB). No depth data or motion capture is used at test time.

\subsection{End-to-End Reinforcement Learning Evaluation}

On the Maniwhere RL suite we benchmark \papername{} against established view-generalization baselines (MV-MWM \cite{seo2023multi}, MoVie \cite{yang2023movie}, SGQN \cite{bertoin2022look}, PIE-G \cite{yuan2022pre}, and SRM \cite{huang2022spectrum}).

To isolate the Maniwhere baseline's dependence on depth, we remove its depth channel and leave the rest unchanged (\textit{Maniwhere (RGB)} in Table~\ref{table:sim-results}). It retains $72.5\%$ and $82.4\%$ on \textit{Lift Cube Dex} and \textit{Button Dex} but drops to $0.0\%$ and $17.3\%$ on \textit{Pick \& Place Dex} and \textit{Close Dex}. Both collapsing tasks require resolving depth-dependent object position under viewpoint shift, the obstacle Maniwhere identifies for monocular RGB~\cite{yuan2024learning}; that robustness comes from the depth channel, not the RGB stream.

The privileged 3D auxiliary loss recovers much of this gap from RGB alone. \papername{} reaches $72.4\%$ on \textit{Pick \& Place Dex} and $92.1\%$ on \textit{Close Dex}, where success depends on the front-to-back hand-lid position under a shifting camera; the absolute-coordinate anchor supplies the metric depth monocular RGB lacks, exceeding RGB-D Maniwhere ($81.5\%$). \textit{Lift Cube Dex} is the exception: fine localization of a small cube favors a spatial feature grid, and global average pooling drops \papername{} to $53.0\%$, below MV-MWM ($78.0\%$) and \textit{Maniwhere (RGB)} ($72.5\%$).

\subsection{Student-Teacher Distillation Evaluation}

To test whether the geometric anchor scales to high-fidelity environments under dense supervision, we evaluate \papername{} using student-teacher distillation across a $140^\circ$ azimuthal training cone ($-110^\circ$ to $30^\circ$), with additional evaluations beyond both boundaries. Relative to the privileged teacher, \papername{} recovers \textbf{61--72\%} of teacher capacity across the training cone, compared with 33--49\% for domain randomization alone, $\leq29\%$ without the auxiliary loss, and $\leq21\%$ without InfoNCE. We further evaluate the distilled policy on physical hardware under uncalibrated camera perturbations in Table.~\ref{table:ablation-results}.

\begin{table}[h]
\centering
\footnotesize
\renewcommand{\arraystretch}{0.92}
\caption{%
  \textbf{Hardware Zero-Shot Grasping.}
  Successes out of 80 trials per viewpoint
  (8 objects $\times$ 10 trials; 480 total per method).
}
\label{table:ablation-results}
\resizebox{\columnwidth}{!}{%
\begin{tabular}{lccccccrr}
\toprule
\multirow{2}{*}{\textbf{Method}}
  & \multicolumn{6}{c}{\textbf{Viewpoint (successes\,/\,80)}}
  & \multirow{2}{*}{\textbf{Total}}
  & \multirow{2}{*}{\textbf{Avg\,\%\,$\uparrow$}} \\
\cmidrule(lr){2-7}
  & $\theta_1$ & $\theta_2$ & $\theta_3$
  & $\theta_4$ & $\theta_5$ & $\theta_6$ & & \\
\midrule
Fixed Camera   &  1 &  0 &  4 &  2 &  0 &  0 &   7/480 &  1.4 \\
DR Only        & 24 & 26 & 24 & 28 & 22 & 21 & 145/480 & 30.2 \\
w/o InfoNCE    & 15 & 15 & 13 & 16 & 16 & 15 &  90/480 & 18.7 \\
w/o Aux.\ Loss &  6 &  8 &  8 &  6 &  5 &  7 &  40/480 &  8.3 \\
\midrule\midrule
\rowcolor{green!15}
\textbf{\papername{} (Ours)}
  & \textbf{60} & \textbf{63} & \textbf{61}
  & \textbf{60} & \textbf{64} & \textbf{60}
  & \textbf{368/480} & \textbf{76.7} \\
\bottomrule
\end{tabular}%
}
\end{table}
\begin{figure}[htbp]
   \centering
   \includegraphics[width=\columnwidth]{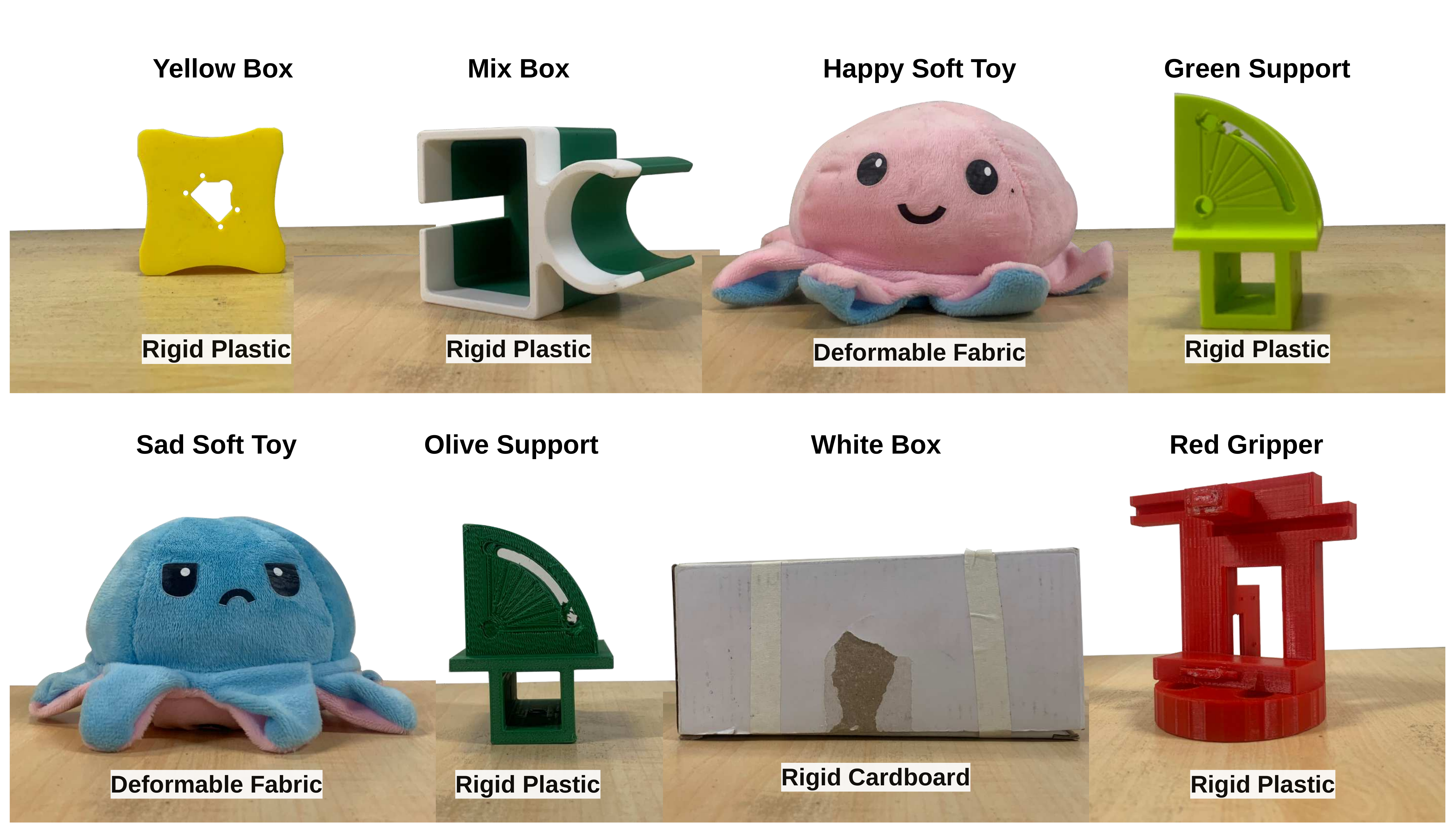}
   \caption{The eight unseen evaluation objects (Object-A--H), spanning rigid plastics, rigid cardboard, and deformable fabrics to test geometric and material generalization.}
   \label{fig:hardware_test_objects}
\end{figure}
\subsection{Visual Feature Probing and 3D Grounding}
\label{subsec:probing}
To test whether the visual backbone encodes metric geometry rather than proprioceptive memorization, we train a linear probe on the frozen 1D visual embedding ($\mathbf{v}_t$), before sensor fusion, to regress 3D object coordinates. Over roughly 690k in-distribution samples the probe reaches $R^2=0.814$ with a mean Euclidean error of 4.73~cm from uncalibrated monocular RGB. A frozen, proprioception-free embedding that decodes linearly to metric coordinates shows the geometry is present in the visual features, not supplied at contact by kinematics.

The 4.73~cm error is open-loop, from a single in-distribution frame without temporal context or proprioception. It is coarse for direct fingertip placement; the closed-loop policy reduces the residual by fusing $\mathbf{v}_t$ with proprioception over time to establish contact. Outside the training cone the probe error rises to $\approx$14.3~cm and the auxiliary prediction degrades (Section~\ref{sec:conclusion}); \papername{} targets calibration-free repositioning within the training distribution, not arbitrary novel-view geometric generalization.

\subsection{Real-World Zero-Shot Deployment}
Hardware evaluation uses a single uncalibrated RealSense D455 repositioned across six azimuthal angles ($\theta_1$ to $\theta_6$) within a $140^\circ$ cone (Figure~\ref{fig:3_panel_deployment}). The test set comprises eight unseen objects varying in geometry and material (Figure~\ref{fig:hardware_test_objects}). We execute 10 trials per object-viewpoint combination. A trial is recorded as a success if the LEAP Hand establishes a multi-fingered grasp and lifts the object completely clear of the tabletop surface.

Success is consistent across viewpoints (75.0--80.0\%, 60--64/80; Table~\ref{table:ablation-results}) and across the eight objects (71.7--83.3\%, 43--50/60 per object; Fig.~\ref{fig:hardware_test_objects}). The hardware study compares matched internal variants of \papername{}; the head-to-head comparison against depth is in simulation (Table~\ref{table:sim-results}). The policy transfers without real-world fine-tuning.

\subsection{Ablation Studies}
We isolate each component against matched baselines, running all four conditions in both paradigms: simulation (Maniwhere RL, Table~\ref{table:sim-results}) and hardware (DextrAH distillation, Table~\ref{table:ablation-results}). \textit{Fixed Camera} uses the canonical view only; \textit{DR Only} adds camera randomization without either representation loss; \textit{w/o Aux} keeps InfoNCE and DR; \textit{w/o InfoNCE} keeps the 3D anchor and DR, resembling a DextrAH-RGB-style aux+DR signal.

\textbf{Simulation (Maniwhere RL).} \textit{Fixed Camera} scores $0\%$ at novel viewpoints; \textit{DR Only} handles translation (Lift Cube $26.2\%$) but collapses on rotation (Close Dex $0\%$). Removing either loss lowers success on every task relative to the full model: \textit{w/o Aux} falls near-zero on three of four tasks, retaining $88.0\%$ on \textit{Button Dex} where a fixed-location press needs no metric anchor; \textit{w/o InfoNCE} caps at $22$--$35\%$.

\textbf{Hardware (DextrAH distillation).} The full model reaches $76.7\%$, above \textit{DR Only} ($30.2\%$), \textit{w/o InfoNCE} ($18.7\%$), \textit{w/o Aux} ($8.3\%$), and \textit{Fixed Camera} ($1.4\%$). Neither loss added to DR exceeds DR alone; only their combination does.

The full objective outperforms every ablation in both paradigms.

\subsection{Deployment Efficiency}

At deployment, our ResNet-18 student runs in half-precision on a consumer-grade NVIDIA RTX 4050 laptop GPU, consuming 1--2~GB VRAM. The end-to-end control step averages \textbf{5.16~ms}; the network forward pass requires 5.65~GFLOPs and \textbf{2.74~ms}. As one heavy-encoder reference point, a DINO ViT-S/16 backbone \cite{caron2021emerging, dosovitskiy2020image} requires 52.59~GFLOPs and 10.79~ms under the same setup.

\subsection{Representation Analysis}

\paragraph{Activation Map Analysis.}

\begin{figure}[H]
   \centering
   \includegraphics[width=\columnwidth]{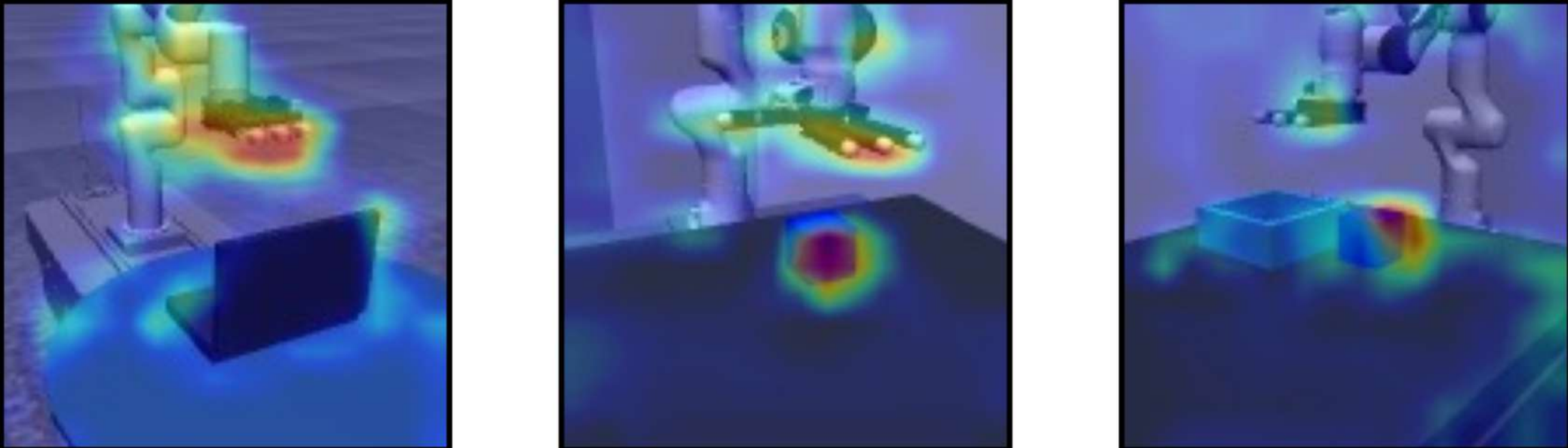}
   \caption{Activation maps extracted from an intermediate residual block before global pooling. Across multiple sampled viewpoints, activations generally appear localized near the finger-object contact areas.}
   \label{fig:attention_map}
\end{figure}

To observe spatial grounding despite Global Average Pooling collapsing the 2D grid, we extract activation maps from an intermediate residual block (Figure~\ref{fig:attention_map}). We adapt Grad-CAM for continuous control by taking gradients of the action magnitude (the $L_2$ norm of the action mean) with respect to these feature maps, highlighting regions that drive the control outputs.

\begin{figure}[htbp]
   \centering
   \includegraphics[width=0.95\linewidth]{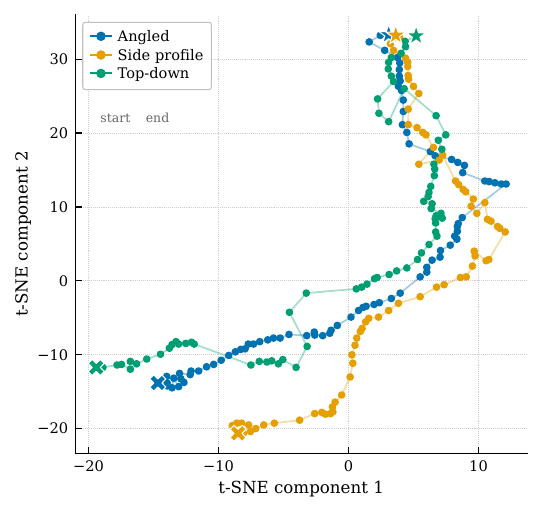}
   \caption{Qualitative t-SNE projection of latent trajectories from three camera viewpoints during a single task execution. Trajectories track together for most of the task and separate near the terminal goal state. t-SNE is non-metric, so distances are indicative only.}
   \label{fig:tsne}

    \vspace{-0.4cm}
\end{figure}

\paragraph{Latent Space (t-SNE) Analysis.}
As a qualitative check, we project the encoder's latent trajectories from three camera angles with t-SNE (Fig.~\ref{fig:tsne}). t-SNE is non-metric and sensitive to its hyperparameters, so we read only coarse structure: trajectories from different views track together through most of the task and separate near the terminal state. The separation is consistent with viewpoint-specific occlusions becoming most pronounced at the goal, though the projection alone does not establish that cause.

\paragraph{Visual Perturbation Analysis.}
To test whether the policies use closed-loop visual feedback rather than memorized open-loop trajectories, we evaluate under spatial masking (random contiguous black/white patches $\Omega$) and pixel noise ($\epsilon \sim \mathcal{N}(0, \sigma^2 I)$ and $\epsilon \sim \mathcal{U}(-a, a)$).

\begin{table}[htbp]
\centering
\footnotesize
\renewcommand{\arraystretch}{1.15}
\setlength{\tabcolsep}{4pt}
\caption{%
  \textbf{Closed-Loop Robustness to Visual Perturbations.}
  Success rate (\%, mean $\pm$ std over 5 seeds; $\uparrow$ higher is better)
  under test-time spatial masking and additive pixel noise on the RGB stream.
  Gradual performance degradation under moderate corruption indicates the policy relies on
  active visual feedback rather than open-loop memorization.
}
\label{tab:robustness}
\resizebox{\columnwidth}{!}{%
\begin{tabular}{lcccc}
\toprule
\multirow{2}{*}{\textbf{Perturbation}}
  & \multicolumn{4}{c}{\textbf{Task Success Rate (\%)} $\uparrow$} \\
\cmidrule(lr){2-5}
  & \textbf{Lift Cube Dex} & \textbf{Pick \& Place Dex}
    & \textbf{Close Dex} & \textbf{Button Dex} \\
\midrule
\rowcolor{gray!12}
None (clean)\textsuperscript{$\dagger$}
    & $53.0 \pm 5.0$  & $72.4 \pm 3.6$
    & $92.1 \pm 5.9$  & $96.0 \pm 2.0$ \\
\midrule
Black patch
    & $23.6 \pm 3.9$  & $13.2 \pm 5.9$
    & $62.0 \pm 10.5$ & $52.4 \pm 7.0$ \\
\addlinespace
Gaussian, $\sigma{=}25$
    & $31.2 \pm 14.7$ & $57.2 \pm 3.0$
    & $30.4 \pm 17.1$ & $62.0 \pm 7.4$ \\
Gaussian, $\sigma{=}50$
    & $20.6 \pm 5.6$  & $37.6 \pm 18.1$
    & $\phantom{0}0.0 \pm 0.0$ & $\phantom{0}0.0 \pm 0.0$ \\
\addlinespace
Uniform, $\pm 50$
    & $45.6 \pm 7.4$  & $54.8 \pm 4.2$
    & $34.8 \pm 16.8$ & $13.6 \pm 19.4$ \\
Uniform, $\pm 75$
    & $\phantom{0}6.1 \pm 2.6$ & $26.4 \pm 6.5$
    & $\phantom{0}0.0 \pm 0.0$ & $\phantom{0}0.0 \pm 0.0$ \\
\bottomrule
\end{tabular}%
}

\parbox{\columnwidth}{\footnotesize\textsuperscript{$\dagger$}\,Clean success (no perturbation), reproduced from Table~\ref{table:sim-results} for reference.}
\end{table}

As reported in Table~\ref{tab:robustness}, performance degrades with increasing noise severity, as expected for active closed-loop visual control. The policy retains task success under moderate noise ($\sigma=25$, $\pm50$ uniform), reflecting the regularization provided by contrastive representation learning.

\section{Conclusion and Limitations}
\label{sec:conclusion}

We introduce \papername{}, an asymmetric representation learning framework for view-invariant dexterous manipulation from monocular RGB. To counter the spatial collapse of a globally pooled contrastive embedding, \papername{} adds a privileged 3D auxiliary anchor during simulated training, giving implicit spatial grounding without depth sensors or 2D warping.

Across reinforcement learning and student-teacher distillation, this architecture mitigates 2D spatial collapse and narrows the gap to depth-reliant systems on contact-rich articulated manipulation while operating from monocular RGB. Physical deployment on a LEAP Hand shows the geometric anchor bridges the sim-to-real gap, enabling zero-shot grasping under continuous camera perturbations.

\textbf{Limitations and Future Work.} \papername{} introduces three key trade-offs. First, compressing inputs into a globally pooled 1D vector discards the 2D spatial grid, slightly degrading tracking for small, low-occlusion objects. Second, implicit calibration relies on the robot's embodiment as a visual reference; total occlusion can cause a loss of absolute scale. Third, regressing a single 3D coordinate limits scalability in clutter. Future work will explore localized, patch-based attention to enhance fine-grained tracking while maintaining global pooling efficiency, as well as expanding the auxiliary head for multiple targets (e.g., via task conditioning or $K \times 3$ regression). We attribute the gains to geometric grounding on the basis of the linear probe and the \textit{w/o Aux} ablation; we do not isolate geometric supervision from a generic dense auxiliary, which a non-geometric control target of matched dimensionality would settle.

\paragraph{Failure Modes.}
We observed two failure modes in real-world deployment. First, a textureless background placed close to the target degrades grasping: global average pooling aggregates features across the frame, so low object-background contrast prevents the network from separating the two. Increasing the object-background distance restored performance. Second, under severely out-of-distribution viewpoints the auxiliary 3D coordinate prediction degrades, producing a geometric mismatch: the LEAP Hand executes kinematically correct grasps in empty air on a plane offset from the target.

\addtolength{\textheight}{-0.35cm}   

\section*{Acknowledgment}
Gemini (Google AI Studio), Claude, and ChatGPT were used to draft and refine text and argument phrasing throughout the paper. Claude was also used to refine an author-made draft of Fig.~\ref{fig:teaser} via its XML source and then manually finalized it. The authors reviewed all content and take full responsibility for it.

\appendix
\label{sec:appendix}
\subsection{Environment and Task Details}
\label{sec:task_setup}
\subsubsection{Tasks}
From the Maniwhere suite~\cite{yuan2024learning} we evaluate four dexterous-hand tasks. \textit{Lift Cube Dex}: grasp a cube and lift it off the table. \textit{Button Dex}: press a button. \textit{Pick \& Place Dex}: grasp a cube and place it into a box. \textit{Close Dex}: close a hinged laptop lid, counted as success below a lid joint angle of $1.7$~rad. For distillation we adapt the DextrAH-RGB pipeline~\cite{singh2025dextrah}, targeting zero-shot object-picking and stable grasping under varying positions.

\subsubsection{Reward Functions}

To ensure fair comparison and reproducibility, we directly adopt the standard dense reward formulations and regularization penalties provided by the respective benchmark suites, Maniwhere \cite{yuan2024learning} and DextrAH-RGB \cite{singh2025dextrah}, without any modifications.

\subsubsection{Domain Randomization}

Both environments heavily randomize physical dynamics and visual properties during training to bridge the sim-to-real gap.

\textbf{Maniwhere Tasks.} To maintain optimization stability, the magnitude of these domain perturbations is governed by an exponential curriculum scheduler ($\gamma=0.995$), which scales baseline robot joint properties (e.g., armature, damping, friction), geometric dimensions (object sizes, table heights), and visual colors by approximately $\pm 10\%$.

\textbf{DextrAH Task.} The DextrAH pipeline employs an Automatic Domain Randomization (ADR) curriculum during privileged teacher training, scaling physics parameters across 50 difficulty levels. The student distillation phase bridges the visual gap with appearance randomization across physical dynamics (object mass, friction, joint PD gains, observation noise), visual materials (HDRI lighting, textures, PBR properties), and image-space augmentations (background swaps, color jitter).

\begin{table}[htbp]
\centering
\footnotesize
\renewcommand{\arraystretch}{0.92}
\caption{%
  \textbf{Key Training Hyperparameters.}
  Privileged teacher (PPO) and visual student (DAgger); the complete
  configuration is released with our code.
}
\label{tab:combined_hyperparameters}
\resizebox{\columnwidth}{!}{%
\begin{tabular}{@{}ll@{\hspace{2.2em}}ll@{}}
\toprule
\multicolumn{2}{c}{\textbf{Teacher (PPO)}} &
\multicolumn{2}{c}{\textbf{Student (DAgger)}} \\
\cmidrule(r){1-2}\cmidrule(l){3-4}
\textbf{Parameter} & \textbf{Value} &
\textbf{Parameter} & \textbf{Value} \\
\midrule
$\gamma$\,/\,$\lambda$   & $0.998$\,/\,$0.95$      & Learning rate      & $1{\times}10^{-4}$ \\
Learning rate            & $3{\times}10^{-4}$      & DAgger $\beta$     & $1.0$ (decay $0.05$) \\
Minibatch                & $16{,}384$              & Contrastive weight & $0.5$ \\
PPO clip $\varepsilon$   & $0.2$                   & LSTM\,/\,MLP       & $512$\,/\,$[512,512,256]$ \\
Policy LSTM\,/\,MLP      & $1024$\,/\,$[512,512]$  & Color jitter       & $\pm50\%$, hue $\pm0.15$ \\
Value LSTM\,/\,MLP       & $2048$\,/\,$[1024,512]$ & BG swap\,/\,blur    & $p{=}0.5$\,/\,$p{=}0.1$ \\
\bottomrule
\end{tabular}%
}
\end{table}
\subsection{Training Details}
\subsubsection{Training Compute}
All policy training is conducted on a single NVIDIA RTX 5090. In Isaac Lab, the privileged teacher is trained across 4{,}096 parallel environments ($\sim$48~h), followed by visual student distillation across 128 environments ($\sim$14~h); MuJoCo RL benchmarks use 256 environments ($\sim$12~h per task).
\subsubsection{Hyperparameters (PPO + DAgger)}
A condensed summary of the critical hyperparameters defining our network architectures, PPO teacher training, and DAgger student distillation is provided in Table~\ref{tab:combined_hyperparameters}.
\bibliographystyle{IEEEtran}
\bibliography{example}

@inproceedings{qin2023dexpoint,
  title={Dexpoint: Generalizable point cloud reinforcement learning for sim-to-real dexterous manipulation},
  author={Qin, Yuzhe and Huang, Binghao and Yin, Zhao-Heng and Su, Hao and Wang, Xiaolong},
  booktitle={Conference on Robot Learning},
  pages={594--605},
  year={2023},
  organization={PMLR}
}

@article{mandlekar2023mimicgen,
  title={Mimicgen: A data generation system for scalable robot learning using human demonstrations},
  author={Mandlekar, Ajay and Nasiriany, Soroush and Wen, Bowen and Akinola, Iretiayo and Narang, Yashraj and Fan, Linxi and Zhu, Yuke and Fox, Dieter},
  journal={arXiv preprint arXiv:2310.17596},
  year={2023}
}

@article{chen2023visual,
  title={Visual dexterity: In-hand reorientation of novel and complex object shapes},
  author={Chen, Tao and Tippur, Megha and Wu, Siyang and Kumar, Vikash and Adelson, Edward and Agrawal, Pulkit},
  journal={Science Robotics},
  volume={8},
  number={84},
  pages={eadc9244},
  year={2023},
  publisher={American Association for the Advancement of Science}
}

@article{yuan2024learning,
  title={Learning to manipulate anywhere: A visual generalizable framework for reinforcement learning},
  author={Yuan, Zhecheng and Wei, Tianming and Cheng, Shuiqi and Zhang, Gu and Chen, Yuanpei and Xu, Huazhe},
  journal={arXiv preprint arXiv:2407.15815},
  year={2024}
}

@article{li2026manivid,
  title={Manivid-3d: Generalizable view-invariant reinforcement learning for robotic manipulation via disentangled 3d representations},
  author={Li, Zheng and Qu, Pei and Jia, Yufei and Zhou, Shihui and Ge, Haizhou and Cao, Jiahang and Zhou, Jinni and Zhou, Guyue and Ma, Jun},
  journal={IEEE Robotics and Automation Letters},
  year={2026},
  publisher={IEEE}
}

@article{yang2023movie,
  title={Movie: Visual model-based policy adaptation for view generalization},
  author={Yang, Sizhe and Ze, Yanjie and Xu, Huazhe},
  journal={Advances in Neural Information Processing Systems},
  volume={36},
  pages={21507--21523},
  year={2023}
}

@inproceedings{pang2025learning,
  title={Learning view-invariant world models for visual robotic manipulation},
  author={Pang, Jing-Cheng and Tang, Nan and Li, Kaiyuan and Tang, Yuting and Cai, Xin-Qiang and Zhang, Zhen-Yu and Niu, Gang and Sugiyama, Masashi and Yu, Yang},
  booktitle={International Conference on Learning Representations},
  volume={2025},
  pages={54853--54876},
  year={2025}
}

@inproceedings{seo2023multi,
  title={Multi-view masked world models for visual robotic manipulation},
  author={Seo, Younggyo and Kim, Junsu and James, Stephen and Lee, Kimin and Shin, Jinwoo and Abbeel, Pieter},
  booktitle={International Conference on Machine Learning},
  pages={30613--30632},
  year={2023},
  organization={PMLR}
}

@inproceedings{seo2023masked,
  title={Masked world models for visual control},
  author={Seo, Younggyo and Hafner, Danijar and Liu, Hao and Liu, Fangchen and James, Stephen and Lee, Kimin and Abbeel, Pieter},
  booktitle={Conference on Robot Learning},
  pages={1332--1344},
  year={2023},
  organization={PMLR}
}

@article{chen2024rovi,
  title={Rovi-aug: Robot and viewpoint augmentation for cross-embodiment robot learning},
  author={Chen, Lawrence Yunliang and Xu, Chenfeng and Dharmarajan, Karthik and Irshad, Muhammad Zubair and Cheng, Richard and Keutzer, Kurt and Tomizuka, Masayoshi and Vuong, Quan and Goldberg, Ken},
  journal={arXiv preprint arXiv:2409.03403},
  year={2024}
}

@article{tian2024view,
  title={View-invariant policy learning via zero-shot novel view synthesis},
  author={Tian, Stephen and Wulfe, Blake and Sargent, Kyle and Liu, Katherine and Zakharov, Sergey and Guizilini, Vitor and Wu, Jiajun},
  journal={arXiv preprint arXiv:2409.03685},
  year={2024}
}

@article{singh2025dextrah,
  title={Dextrah-rgb: Visuomotor policies to grasp anything with dexterous hands, 2025},
  author={Singh, Ritvik and Allshire, Arthur and Handa, Ankur and Ratliff, Nathan and Van Wyk, Karl},
  journal={URL https://arxiv.org/abs/2412.01791},
  volume={9},
  year={2025}
}

@article{bertoin2022look,
  title={Look where you look! saliency-guided q-networks for generalization in visual reinforcement learning},
  author={Bertoin, David and Zouitine, Adil and Zouitine, Mehdi and Rachelson, Emmanuel},
  journal={Advances in neural information processing systems},
  volume={35},
  pages={30693--30706},
  year={2022}
}

@article{oord2018representation,
  title={Representation learning with contrastive predictive coding},
  author={Oord, Aaron van den and Li, Yazhe and Vinyals, Oriol},
  journal={arXiv preprint arXiv:1807.03748},
  year={2018}
}

@article{huang2022spectrum,
  title={Spectrum random masking for generalization in image-based reinforcement learning},
  author={Huang, Yangru and Peng, Peixi and Zhao, Yifan and Chen, Guangyao and Tian, Yonghong},
  journal={Advances in Neural Information Processing Systems},
  volume={35},
  pages={20393--20406},
  year={2022}
}

@article{yuan2022pre,
  title={Pre-trained image encoder for generalizable visual reinforcement learning},
  author={Yuan, Zhecheng and Xue, Zhengrong and Yuan, Bo and Wang, Xueqian and Wu, Yi and Gao, Yang and Xu, Huazhe},
  journal={Advances in Neural Information Processing Systems},
  volume={35},
  pages={13022--13037},
  year={2022}
}

@inproceedings{he2016deep,
  title={Deep residual learning for image recognition},
  author={He, Kaiming and Zhang, Xiangyu and Ren, Shaoqing and Sun, Jian},
  booktitle={Proceedings of the IEEE conference on computer vision and pattern recognition},
  pages={770--778},
  year={2016}
}

@inproceedings{ross2011reduction,
  title={A reduction of imitation learning and structured prediction to no-regret online learning},
  author={Ross, St{\'e}phane and Gordon, Geoffrey and Bagnell, Drew},
  booktitle={Proceedings of the $14^{th}$ International Conference on Artificial Intelligence and Statistics},
  pages={627--635},
  year={2011},
  organization={JMLR Workshop and Conference Proceedings}
}

@article{jiang2025you,
  title={Do you know where your camera is? view-invariant policy learning with camera conditioning},
  author={Jiang, Tianchong and Ji, Jingtian and Tan, Xiangshan and Fang, Jiading and Bhattad, Anand and Guizilini, Vitor and Walter, Matthew R},
  journal={arXiv preprint arXiv:2510.02268},
  year={2025}
}

@inproceedings{walke2023bridgedata,
  title={Bridgedata v2: A dataset for robot learning at scale},
  author={Walke, Homer Rich and Black, Kevin and Zhao, Tony Z and Vuong, Quan and Zheng, Chongyi and Hansen-Estruch, Philippe and He, Andre Wang and Myers, Vivek and Kim, Moo Jin and Du, Max and others},
  booktitle={Conference on robot learning},
  pages={1723--1736},
  year={2023},
  organization={PMLR}
}

@article{hochreiter1997long,
  title={Long short-term memory},
  author={Hochreiter, Sepp and Schmidhuber, J{\"u}rgen},
  journal={Neural computation},
  volume={9},
  number={8},
  pages={1735--1780},
  year={1997},
  publisher={MIT press}
}

@InProceedings{Xie_2021_CVPR,
    author    = {Xie, Zhenda and Lin, Yutong and Zhang, Zheng and Cao, Yue and Lin, Stephen and Hu, Han},
    title     = {Propagate Yourself: Exploring Pixel-Level Consistency for Unsupervised Visual Representation Learning},
    booktitle = {Proceedings of the IEEE/CVF Conference on Computer Vision and Pattern Recognition (CVPR)},
    month     = {June},
    year      = {2021},
    pages     = {16684-16693}
}

@INPROCEEDINGS{9578497,
  author={Wang, Xinlong and Zhang, Rufeng and Shen, Chunhua and Kong, Tao and Li, Lei},
  booktitle={2021 IEEE/CVF Conference on Computer Vision and Pattern Recognition (CVPR)}, 
  title={Dense Contrastive Learning for Self-Supervised Visual Pre-Training}, 
  year={2021},
  volume={},
  number={},
  pages={3023-3032},
  doi={10.1109/CVPR46437.2021.00304}}

@article{andrychowicz2020learning,
  title={Learning dexterous in-hand manipulation},
  author={Andrychowicz, OpenAI: Marcin and Baker, Bowen and Chociej, Maciek and Jozefowicz, Rafal and McGrew, Bob and Pachocki, Jakub and Petron, Arthur and Plappert, Matthias and Powell, Glenn and Ray, Alex and others},
  journal={The International Journal of Robotics Research},
  volume={39},
  number={1},
  pages={3--20},
  year={2020},
  publisher={SAGE Publications Sage UK: London, England}
}

@inproceedings{keselman2017intel,
  title={Intel (r) realsense (tm) stereoscopic depth cameras},
  author={Keselman, Leonid and Woodfill, John Iselin and Grunnet-Jepsen, Anders and Bhowmik, Achintya},
  booktitle={2017 IEEE conference on computer vision and pattern recognition workshops (CVPRW)},
  pages={1267--1276},
  year={2017},
  organization={IEEE}
}

@article{wang2024dexcap,
  title={Dexcap: Scalable and portable mocap data collection system for dexterous manipulation},
  author={Wang, Chen and Shi, Haochen and Wang, Weizhuo and Zhang, Ruohan and Fei-Fei, Li and Liu, C Karen},
  journal={arXiv preprint arXiv:2403.07788},
  year={2024}
}

@article{mittal2025isaac,
  title={Isaac lab: A gpu-accelerated simulation framework for multi-modal robot learning},
  author={Mittal, Mayank and Roth, Pascal and Tigue, James and Richard, Antoine and Zhang, Octi and Du, Peter and Serrano-Munoz, Antonio and Yao, Xinjie and Zurbr{\"u}gg, Ren{\'e} and Rudin, Nikita and others},
  journal={arXiv preprint arXiv:2511.04831},
  year={2025}
}

@article{shaw2023leap,
  title={Leap hand: Low-cost, efficient, and anthropomorphic hand for robot learning},
  author={Shaw, Kenneth and Agarwal, Ananye and Pathak, Deepak},
  journal={arXiv preprint arXiv:2309.06440},
  year={2023}
}

@inproceedings{caron2021emerging,
  title={Emerging properties in self-supervised vision transformers},
  author={Caron, Mathilde and Touvron, Hugo and Misra, Ishan and J{\'e}gou, Herv{\'e} and Mairal, Julien and Bojanowski, Piotr and Joulin, Armand},
  booktitle={2021 IEEE/CVF international conference on computer vision (ICCV)},
  pages={9630--9640},
  year={2021},
  organization={IEEE}
}

@article{dosovitskiy2020image,
  title={An image is worth 16x16 words: Transformers for image recognition at scale},
  author={Dosovitskiy, Alexey and Beyer, Lucas and Kolesnikov, Alexander and Weissenborn, Dirk and Zhai, Xiaohua and Unterthiner, Thomas and Dehghani, Mostafa and Minderer, Matthias and Heigold, Georg and Gelly, Sylvain and others},
  journal={arXiv preprint arXiv:2010.11929},
  year={2020}
}
\end{document}